\documentclass[letterpaper]{article} 
\usepackage{aaai24}  
\usepackage{times}  
\usepackage{helvet}  
\usepackage{courier}  
\usepackage[hyphens]{url}  
\usepackage{graphicx} 
\usepackage{natbib}  
\usepackage{caption} 
\usepackage{algorithm}
\usepackage{algorithmic}

\usepackage{newfloat}
\usepackage{listings}

\usepackage{times}
\usepackage{epsfig}
\usepackage{graphicx}
\usepackage{amsmath}
\usepackage{amssymb}
\usepackage{pifont}
\usepackage{booktabs}
\usepackage{arydshln}

\usepackage{colortbl}
\usepackage{caption}
\usepackage{subcaption}
\usepackage{multirow}
\usepackage{rotfloat}
\usepackage{capt-of}
\usepackage{longtable}
\usepackage{diagbox}
\usepackage{makecell}
\usepackage{enumitem}

\newcommand{\cmark}{\ding{51}}%
\newcommand{\xmark}{\ding{55}}%

\DeclareCaptionStyle{ruled}{labelfont=normalfont,labelsep=colon,strut=off} 
\floatstyle{ruled}
\newfloat{listing}{tb}{lst}{}
\floatname{listing}{Listing}
\title{DeepAccident: A Motion and Accident Prediction Benchmark for V2X Autonomous Driving}
\author{
    Tianqi Wang\textsuperscript{\rm 1}, Sukmin Kim\textsuperscript{\rm 1},
    Ji Wenxuan\textsuperscript{\rm 1}, Enze Xie\textsuperscript{\rm 2 *},\\
    Chongjian Ge\textsuperscript{\rm 1}, Junsong Chen\textsuperscript{\rm 3},
    Zhenguo Li\textsuperscript{\rm 2}, Ping Luo\textsuperscript{\rm 1 *}\\
}
\affiliations{
    \textsuperscript{\rm 1}The University of Hong Kong\\
    \textsuperscript{\rm 2}Huawei Noah’s Ark Lab\\
    \textsuperscript{\rm 3}Dalian University of Technology
}

\begin{document}

\makeatletter
\twocolumn[{
  \renewcommand\twocolumn[1][]{#1}
  \@maketitle
}]
\makeatother

\begin{figure*}[t!]
\centering
\includegraphics[width=1.0\textwidth]{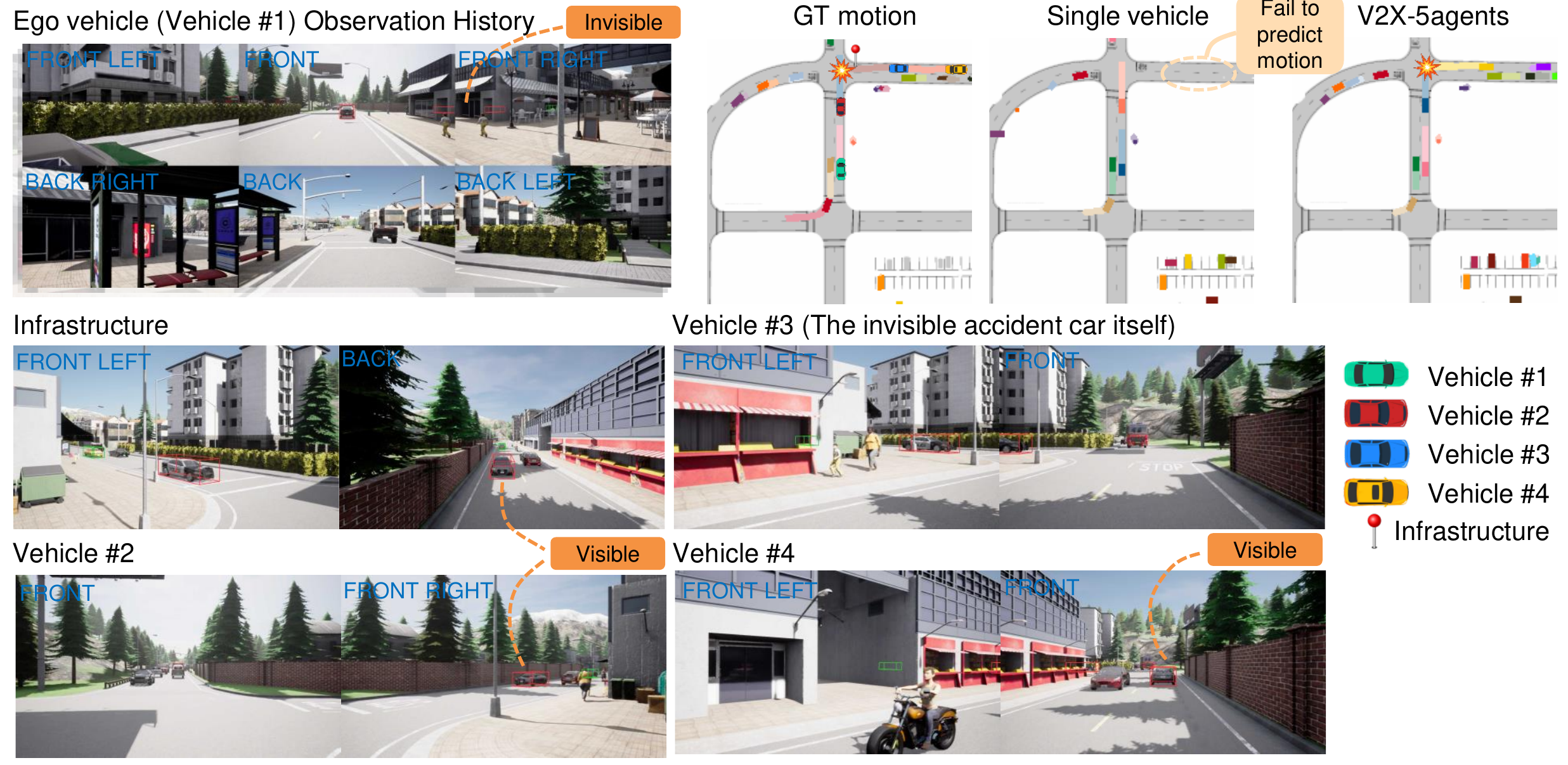}
\caption{Illustration of our proposed end-to-end motion and accident prediction task. Given the history camera observations, the single vehicle model (vehicle \# 1) fails to predict any motion or accident on the forward right side due to occlusion from buildings. In contrast, the V2X model communicates with other vehicles and infrastructure, thereby successfully anticipating the upcoming accident. The red and green bounding boxes in the images, respectively, represent the colliding vehicles and the other V2X vehicles behind them.}
\label{fig:illustrate the v2x accident prediction task}
\end{figure*}

\newcommand\blfootnote[1]{
    \begingroup
    \renewcommand\thefootnote{}\footnote{#1}
    \addtocounter{footnote}{-1}
    \endgroup
}
\blfootnote{$*$ Corresponding authors.}

\begin{abstract}
Safety is the primary priority of autonomous driving. Nevertheless, no published dataset currently supports the direct and explainable safety evaluation for autonomous driving. In this work, we propose DeepAccident, a large-scale dataset generated via a realistic simulator containing diverse accident scenarios that frequently occur in real-world driving. The proposed DeepAccident dataset includes 57K annotated frames and 285K annotated samples, approximately 7 times more than the large-scale nuScenes dataset with 40k annotated samples. In addition, we propose a new task, end-to-end motion and accident prediction, which can be used to directly evaluate the accident prediction ability for different autonomous driving algorithms. Furthermore, for each scenario, we set four vehicles along with one infrastructure to record data, thus providing diverse viewpoints for accident scenarios and enabling V2X (vehicle-to-everything) research on perception and prediction tasks. Finally, we present a baseline V2X model named V2XFormer that demonstrates superior performance for motion and accident prediction and 3D object detection compared to the single-vehicle model.
\end{abstract}

\section{Introduction}
In recent years, single-vehicle autonomous driving has achieved significant progress owing to the well-established datasets for autonomous driving, such as  KITTI~\cite{KITTI}, nuScenes~\cite{nuscenes2020cvpr}, Waymo~\cite{Waymo}, etc. Using those datasets, researchers have proposed various representative algorithms for different downstream tasks, including perception \cite{pointpillar,pvrcnn,bevdet,fasetbev} and prediction \cite{fiery,beverse}.

Nevertheless, single-vehicle autonomous driving suffers from performance degradation in distant or occluded areas due to poor or partial visibility of raw sensors.
One possible solution is to seek the help of vehicle-to-everything (V2X) communication technology which can provide a complementary perception range or enhanced visibility for the ego vehicle. Based on the additional information source, V2X communication can be further categorized as vehicle-to-vehicle (V2V) and vehicle-to-infrastructure (V2I).
Most of the existing V2X datasets \cite{OPV2V,v2x-sim,DAIR-V2X} support perception tasks but ignore the critical motion prediction task. The only V2X dataset that supports motion prediction is V2X-seq \cite{V2X-seq}, that is recently released, but it requires ground truth vehicle positions, map topology, and traffic light status as inputs, which is impractical for real-world autonomous driving.
Moreover, mainstream datasets lack an essential attribute for evaluating the safety of autonomous driving: the inclusion of safety-critical scenarios, such as collision accidents. 
Existing accident datasets \cite{VIENA2, crash_to_not_crash, TAD} suffer from limitations such as low-resolution images captured from a single forward-facing camera and coarse accident annotation labels.

We propose the DeepAccident dataset, the first V2X autonomous driving dataset supporting end-to-end motion and accident prediction, and various perception tasks. Using the CARLA simulator \cite{carla}, we reconstructed diverse real-world driving accidents according to NHTSA pre-crash reports \cite{pre-crash}. For each scenario, four vehicles and one infrastructure are designed to collect full sets of sensor data, including multi-view cameras and LiDAR, with labels for perception and prediction tasks. This setup fills the gap of a lack of safety-critical scenarios in existing V2X datasets. 
Additionally, we introduce a new end-to-end accident prediction task to predict collision accidents' occurrence, timing, location, and involved vehicles or pedestrians. An illustration of this task is shown in Figure \ref{fig:illustrate the v2x accident prediction task}. Lastly, we propose a V2X model named V2XFormer, which demonstrates superior performance compared to the single-vehicle model on the DeepAccident dataset.

Our main contributions can be summarized in three-fold: 
(i) DeepAccident, the first V2X dataset and benchmark that contains diverse collision accidents,
(ii) a new task named end-to-end accident prediction that predicts the occurrence of collision accidents and their specific timing, location, and vehicles or pedestrians involved, 
and (iii) a V2X model named V2XFormer for both perception and prediction tasks to serve as a baseline for further research.

\section{Related Work}
\paragraph{V2X datasets for autonomous driving.} 
The existing V2X datasets primarily focus on perception tasks, including simulator-generated OPV2V \cite{OPV2V} and V2X-Sim \cite{v2x-sim}, and real-world datasets  DAIR-V2X \cite{DAIR-V2X} and V2X-seq \cite{V2X-seq}. V2X-seq is currently the only available dataset that supports the motion prediction task. However, it only supports the traditional motion task, which assumes perfect perception and takes ground truth vehicle locations, map topology, and traffic light status as inputs. Alternatively, end-to-end motion prediction, which takes the raw sensor as input and generates the motion prediction results, has aroused significant research interest recently \cite{fiery,beverse} due to the potential to extract more semantics from the raw sensor and the time efficiency. In comparison, our proposed DeepAccident dataset provides multi-view camera and LiDAR sensor data and supports all common perception tasks and end-to-end motion and accident prediction task refer to Table \ref{tab:accident_comparison}. 
Moreover, DeepAccident also has the largest scale compared with existing datasets according to Table \ref{tab:dataset scale comparison}.

\begin{table*}[t!]
    \begin{center}
    \small
    \renewcommand{\tabcolsep}{2mm}
        \begin{tabular}{l|c|c|c c|c c c c|c}
            \toprule[0.3mm]
            \multirow{2}{*}{Dataset} & \multirow{2}{*}{Scenario} & \multirow{2}{*}{Source} & \multicolumn{2}{c|}{Sensor} & \multicolumn{4}{c|}{Tasks} & \multirow{2}{*}{Accident}\\
             & & & MTV Cameras & LiDAR & Det. & Track. & Seg. & Mot.\\
            \midrule
            nuScenes & single & real-world & \cmark & \cmark & \cmark & \cmark & \cmark & $\triangle$ & \xmark\\
            Waymo & single & real-world & \cmark & \cmark & \cmark & \cmark & \xmark & $\triangle$ & \xmark\\ 
            KITTI & single & real-world & \xmark & \cmark & \cmark & \cmark & \xmark & $\triangle$ & \xmark\\ 
            \midrule
            OPV2V & V2V & simulator & \cmark & \cmark & \cmark & \cmark & \xmark & $\triangle$ & \xmark\\
            V2X-Sim & V2V\&V2I & simulator & \cmark & \cmark & \cmark & \cmark & \cmark & $\triangle$ & \xmark\\ 
            DAIR-V2X & V2I & real-world & \xmark & \cmark & \cmark & \xmark & \xmark & \xmark & \xmark\\
            V2X-seq/Perception & V2I & real-world & \xmark & \cmark & \cmark & \cmark & \xmark & $\triangle$ & \xmark\\ 
            V2X-Seq/Forecasting & V2I & real-world & \xmark & \cmark & \cmark & \cmark & \xmark & \xmark & \xmark\\ 
            \midrule
            VIENA$^2$ & single & simulator & \xmark & \xmark & \xmark & \xmark & \xmark & \xmark & \cmark\\
            GTACrash & single & simulator & \xmark & \xmark & \xmark & \xmark & \xmark & \xmark & \cmark\\             
            YoutubeCrash  & single & real-world & \xmark & \xmark & \xmark & \xmark & \xmark & \xmark & \cmark\\
            TAD & single & real-world  & \xmark & \xmark & \xmark & \xmark & \xmark & \xmark & \cmark\\
            \midrule
            DeepAccident (\textbf{ours}) & V2V\&V2I & simulator & \cmark & \cmark & \cmark & \cmark & \cmark & \cmark & \cmark\\
            
            \bottomrule[0.3mm]
        \end{tabular}
    \caption{Attribute comparison of existing autonomous driving datasets. The Mot. task listed in Tasks represents end-to-end motion prediction, and the symbol $\triangle$ indicates that the required ground truth motion labels are not officially provided but can be obtained via manipulation of the original sequential labels.}
    \label{tab:accident_comparison}
    \end{center}
\end{table*}

\begin{table*}[t!]
    \begin{center}
    \small
    \setlength{\tabcolsep}{2pt}
        \begin{tabular}{c|c c c c c c c}
            \toprule[0.3mm]
            & KITTI & nuScenes & Waymo & OPV2V & V2X-Sim & V2X-seq & DeepAccident (\textbf{ours})\\
            \midrule
            \# of annotated samples & 15K & 40K & 230K & 33K & 47K & 36K & \textbf{285K}\\
            \# of annotated V2X frames& 0 & 0 & 0 & 11K	& 10K	& 18K & \textbf{57K}\\
            annotation frequency (Hz) & 1 & 2 & 10 & 10 & 5 & 10 & \textbf{10}\\
            \bottomrule[0.3mm]
        \end{tabular}
    \caption{Scale comparison of existing autonomous driving datasets to the proposed DeepAccident.} 
    \label{tab:dataset scale comparison}
    \end{center}
\end{table*}

\noindent \textbf{Accident datasets for autonomous driving.} 
Currently, there are few existing accident datasets which only operate within a single vehicle or single infrastructure setting. VIENA$^2$ \cite{VIENA2} and GTACrash \cite{crash_to_not_crash} create collisions in the GTA V video game by manually driving or randomly losing control, thus having limited accident diversity and realism. YoutubeCrash \cite{crash_to_not_crash} and TAD \cite{TAD} respectively utilize real-world collision video clips captured from vehicle forward cameras or infrastructure surveillance cameras. All these datasets only contain low-resolution forward camera images and oversimplify the accident prediction task as a classification or 2D dangerous vehicle detection task, which is challenging to interpret or use for subsequent planning module in autonomous driving.
In contrast, our proposed DeepAccident dataset provides fully detailed accident labels, such as the accident vehicle ids and their future colliding trajectories in the V2X scenario.

\noindent \textbf{Accident scenario generation.} 
The generation of accident scenarios can be categorized as optimization-based and knowledge-based. AdvSim \cite{AdvSim} and STRIVE \cite{STRIVE} belong to the former and generate the perturbed adversary trajectories for other vehicles to attack the fixed ego planner. AdvSim selects adverse vehicles beforehand and optimizes their action profiles with the kinematics bicycle model and black-box optimizations. STRIVE represents the traffic motion as a learned latent vector and leverage gradient-based optimization to optimize the latent vector. However, the optimization-based methods can be prohibitively time-consuming for generating large-scale accident datasets. For instance, generating a single 10-agent 8s scenario takes 6-7 minutes for STRIVE, even several hours for AdvSim. Besides, the optimization-based methods can generate implausible or unrealistic scenarios and thus need additional filtering steps, which adds more computational overhead. Our accident generation method is mostly similar to CARLA's ScenarioRunner and belongs to the knowledge-based method. It uses rules for adverse agent behaviors and generates corresponding trajectories. 
The proposed DeepAccident dataset focus on the accidents that happen in intersections. To increase the trajectory diversity, we randomly choose the starting positions, destination directions, and the maximum speed of ego and adverse vehicles.

\section{DeepAccident Dataset}

\subsection{Dataset Generation}
\begin{figure*}[t!]
  \centering
  \includegraphics[width=1.0\textwidth]{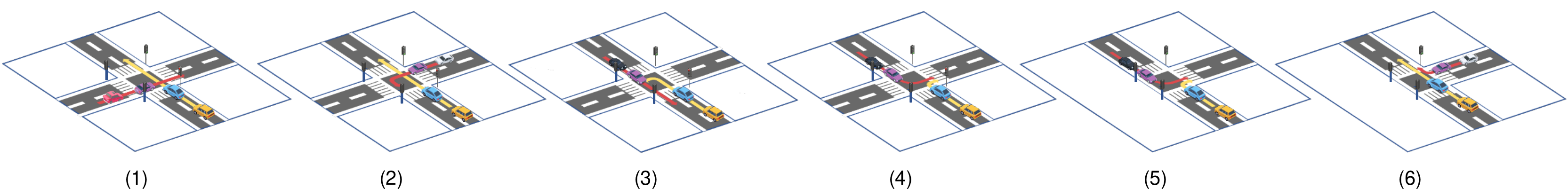}
  \caption{Designed accident scenarios in DeepAccident across signalized intersections and unsignalized intersections. Each scenario involves two colliding vehicles with overlapping trajectories and two following vehicles. The designed scenarios include: (1) running against a red light at four-way intersections, (2) left turn against a red light at four-way intersections, (3) unprotected left turn at four-way intersections, (4) right turn against left turn at four-way intersections, (5) right turn against left turn at three-way intersections (6) go straight against right turn at three-way intersections in signalized cases. In unsignalized cases, the designed overlapping trajectories are the same, but there are no traffic lights to affect the vehicle behaviors.}
  \label{fig:accident types}
\end{figure*}

The accident scenarios in DeepAccident are designed following the pre-crash report by NHTSA, in which various types of collision accidents are reported from real-world crash data. We design 12 types of accident scenarios that happen in intersections in DeepAccident, as shown in Figure \ref{fig:accident types}.
Our designed accident scenarios generally involve two vehicles with overlapped planned trajectories at signalized and unsignalized intersections. In addition to the two accident vehicles, we spawn two more vehicles, each following behind one of the accident vehicles, to capture diverse viewpoints of the same scene (See Figure \ref{fig:accident types}). Furthermore, full sets of sensors are installed on all four vehicles, and annotated labels are saved independently. Additionally, a full stack of sensors is installed facing toward the intersection on the infrastructure side, resulting in data from four vehicles and one infrastructure of the same scene to support V2X research. 
The detailed sensor configuration and an illustration example of our V2X setting are provided in the appendix.

\begin{figure*}[t!]
  \centering
  \includegraphics[width=1.0\textwidth]{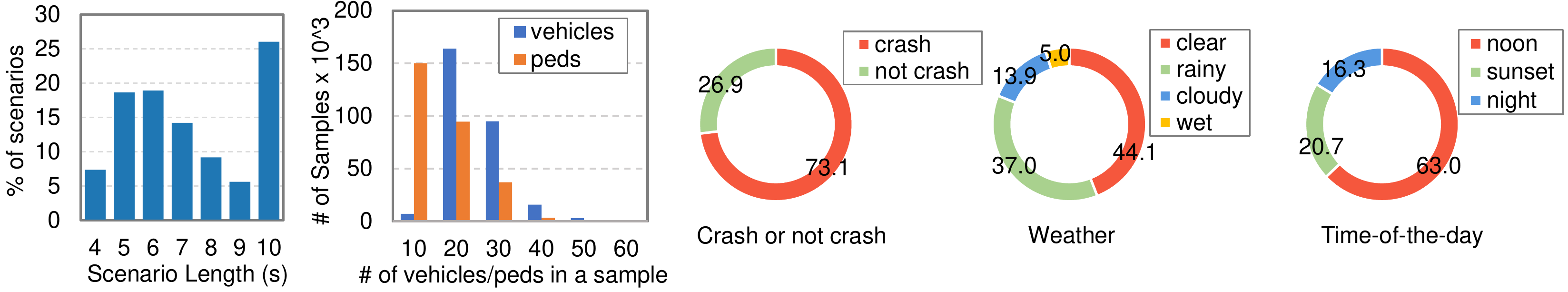}
  \caption{Distribution of the proposed DeepAccident dataset}
  \label{fig:scene statistics}
\end{figure*}

\noindent \textbf{Accident generation details.} 
For the two accident vehicles that we choose to collide with each other, we first calculate the intersection point of their planned trajectories and then perturb their initial positions to arrive at a similar time by dividing the arriving distance by the randomly chosen maximum speeds. The two vehicles designed to follow the accident vehicles have the same trajectories as the accident vehicles.
Note that accident vehicles will also react to the surrounding traffic using rule-based controllers when necessary, thus altering their arrival times at the accident site.
As a result, these accident vehicles may collide at diverse relative positions or angles and could potentially collide with other vehicles to increase trajectory diversity. Despite not being specifically designed, other accident scenarios occurred during data collection, such as vehicles hitting crossing pedestrians and colliding with lane-changing vehicles. These are included in the proposed DeepAccident dataset, with plans to incorporate more scenarios in future versions. Normal scenarios without collisions are also included in DeepAccident to enhance motion diversity.
Each scenario will terminate when a collision occurs, the ego vehicle completes its planned trajectory, or the scenario time exceeds 10 seconds.

\subsection{Dataset Statistics}
\noindent \textbf{Scenario distribution.} 
During data collection, various random factors such as number of surrounding vehicles and pedestrians, weather, and time-of-day were applied to enhance diversity. As depicted in Figure \ref{fig:scene statistics}, the DeepAccident dataset demonstrates significant diversity. 
Detailed statistics are available in the appendix.

\noindent \textbf{Supported tasks.} 
DeepAccident supports various perception tasks such as 3D object detection, tracking, BEV semantic segmentation, and prediction tasks like motion prediction and accident prediction with detailed accident labels. All these tasks can be achieved within V2X settings in DeepAccident, thus stimulating more V2X research.

\noindent \textbf{Dataset size.} 
 The proposed DeepAccident comprises a total of 285k annotated samples and 57k annotated V2X frames at a frequency of 10 Hz. Besides, we split the data with a ratio of 0.7, 0.15, and 0.15 for training, validation, and testing splits, resulting in 203k, 41k, and 41k samples, respectively.

\section{End-to-End Motion and Accident Prediction}
For this task, we select the camera-based setting to utilize multi-view camera image streams as inputs for generating motion prediction results for the entire scene. These motion prediction results are then post-processed to determine the occurrence of the accident and the accident vehicle ids, accident positions as well as timing (see Figure \ref{fig:illustrate the v2x accident prediction task}).


\subsection{Network Structure}

\begin{figure*}[t!]
  \centering
  \includegraphics[width=0.98\textwidth]{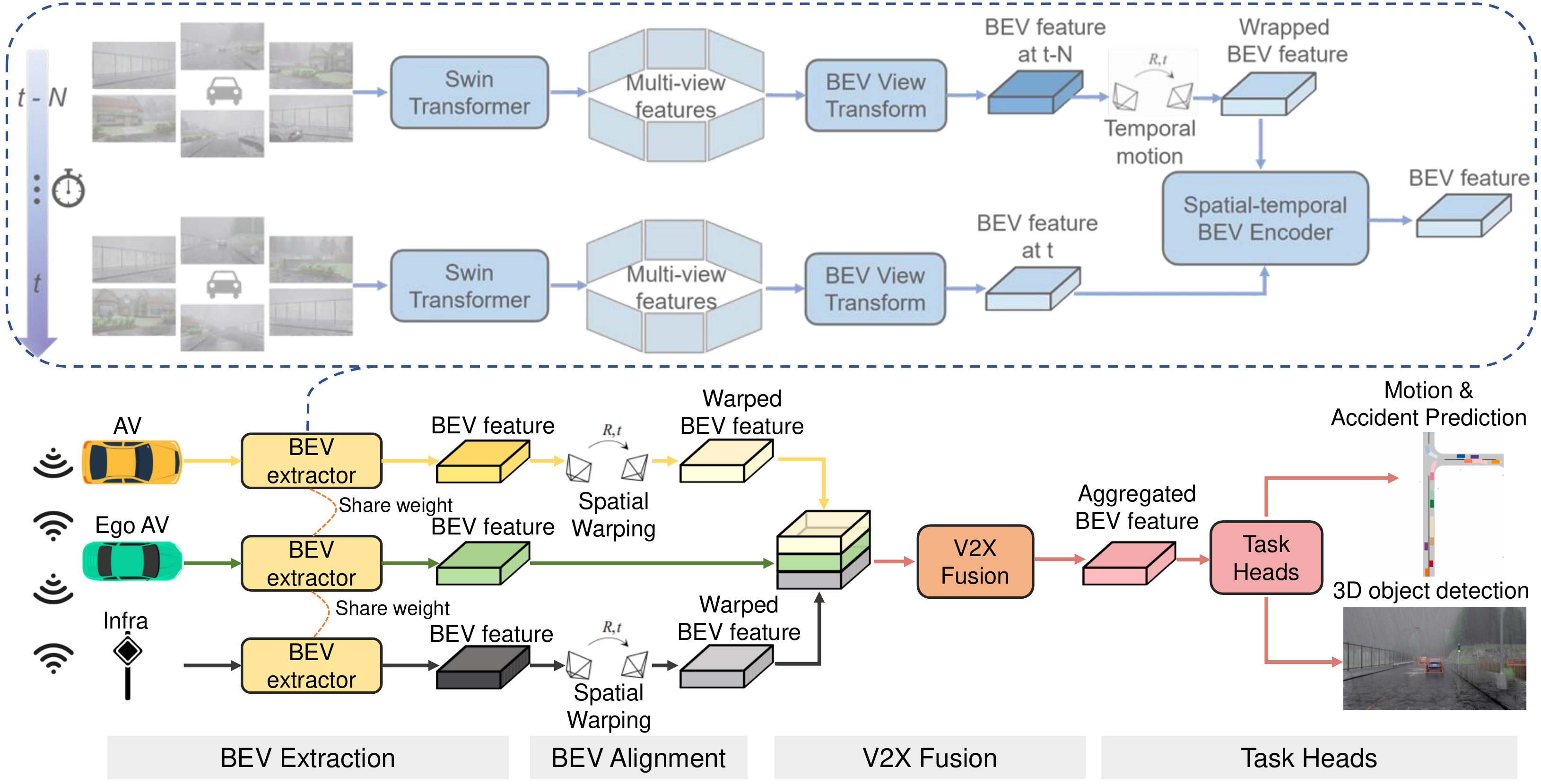}
   \caption{Network details of the proposed V2XFormer. We use the three-V2X-agent setting consisting of ego AV, AV, and Infra for illustration. V2X agents in V2XFormer utilize a shared-weight BEV extractor to extract BEV features based on multi-view camera observation history within the previous N frames.}
  \label{fig:V2XFormer}
\end{figure*}

 We choose BEVerse \cite{beverse} as our baseline single-vehicle model due to its support for end-to-end motion prediction. For the V2X setting, we propose a simple yet effective V2X model named V2XFormer due to using SwinTransformer \cite{swin-transformer} as the image feature backbone. As shown in Figure \ref{fig:V2XFormer}, V2XFormer shares the same BEV feature extractor as the single-vehicle model such that each V2X agent would extract a BEV feature centered at its own coordinate system. These BEV features are then spatially wrapped to the ego vehicle coordinate system to concatenate with the ego vehicle BEV feature. To fuse the concatenated BEV features, we utilize the fusion modules of the state-of-the-art V2X methods. The aggregated BEV feature is then fed into the task heads to generate the motion prediction and 3D object detection results. In addition, given the fact that the ego vehicle itself can cause an accident with other vehicles or pedestrians, we also require the network to jointly predict the ego vehicle's future motion. 

\subsection{Accident Prediction} \label{subsec: accident prediction}
\noindent \textbf{Post-processing for accident prediction.} We post-process the motion prediction results frame-wise to check the occurrence of accident. The performance of accident prediction can be viewed as a safety metric for autonomous driving. From the motion prediction results, which consist of several BEV outputs, including centerness, segmentation, offset, and future flow, we can combine them to get the BEV instance segmentation results like the ones shown in Figure \ref{fig:illustrate the v2x accident prediction task} for the current moment as well as the future period. For each timestamp, we can approximate the BEV segmentation results for each object as polygons and then find the polygons with the closest distance and store the object ids and positions to represent the accident candidates for this timestamp. By looking for the timestamp with the closest object distance, we determine whether an accident occurred and provide labels regarding the colliding object ids and positions, and the collision timestamp. In our experiment, we set a threshold for a dangerous distance as 1.0 meters.

\noindent \textbf{Accident prediction accuracy.} To evaluate the accuracy of accident prediction compared to ground truth accident information, the same post-process steps are applied to the ground truth motion to determine the occurrence of future accidents. A true positive prediction is when : (i) both the prediction and ground truth indicate the occurrence of an accident, and (ii) the total position difference of the colliding agents between the prediction and the ground truth is less than a threshold. Based on this, we propose a new evaluation metric named Accident Prediction Accuracy (APA) in Equation \ref{eq:accident acc} where we calculate the average accident prediction accuracy over a set of position difference thresholds of $\mathbb{D} = $ \{5,10,15\} meters:

\begin{equation}
\label{eq:accident acc}
\mathrm{APA} = \frac{1}{\mathbb{D}} \sum_{d \in \mathbb{D}} \frac{|TP|_d}{|TP|_d + \frac{1}{2}|FP|_d + \frac{1}{2}|FN|_d}
\end{equation}

\noindent \textbf{True Positive metrics.} In addition to the APA, we calculate several \emph{True Positive} metrics (TP metrics) for true positive accident predictions to provide more detailed performance interpretation. This includes the error terms for \emph{IDs}, \emph{positions} and \emph{time} between the ground truth accident and the predicted accident. For TP metrics calculation, we set the position difference threshold to 10 meters when deciding the true positive predictions. As for ID error, if the predicted accident objects' ids are the same as the ground truth's, then it equals to zero, otherwise equals to one. For position and time error, we present them using their native units (\emph{meters} and \emph{seconds}) and calculate the absolute difference compared to the ground truth. For each TP metric, we calculate the average value over all true positive predictions.

\section{Experiment}
\noindent \textbf{Evaluated tasks.} To show the usefulness of our proposed DeepAccident dataset as a V2X motion and accident prediction benchmark, we focus on the end-to-end motion and accident prediction task and choose the camera-based setting. Besides, we train another 3D object detection head with the motion head to simultaneously compare the perception ability between the V2X models and the single-vehicle model.

\noindent \textbf{Experiment settings.} We use the settings for motion prediction in BEVerse \cite{beverse} and FIERY \cite{fiery} as our default experiment settings. For 3D object detection, the BEV ranges are [-51.2m, 51.2m] for both X-axis and Y-axis with a 0.8m interval, while for motion prediction, the ranges are [-50.0m, 50.0m] with a 0.5m interval. The models use 1 second of past observations to predict 2 seconds into the future, corresponding to a temporal context of 3 past frames including the current frame and 4 future frames at 2Hz. We choose BEVerse-tiny as the single-vehicle model. For training, we train the models on the training split of DeepAccident for 20 epochs. As for evaluation, we randomly sample five BEV features from the learned motion Gaussian distribution along with the mean of this learned distribution to generate six different motion prediction results. Only the motion prediction result obtained from the mean vector of the learned Gaussian distribution is used to assess motion prediction performance. For accident prediction, we consider a prediction indicating the occurrence of an accident when any of the sampled motion predictions is analyzed to cause a collision accident, prioritizing safety.

We report the performance on DeepAccident's validation split in the following sections and include results on the testing split in the appendix. 
We start by comparing different V2X fusion modules and choosing the optimal one. After that, we compare the overall performance of V2X models with different agent configurations to the single-vehicle model and provide further ablation analysis that considers accident visibility, longer prediction horizon, and robustness on pose error and latency. Additionally, we conduct experiments on nuScenes to validate the trained models' real-world generalization ability.

\noindent \textbf{Evaluation metrics.} We use mIOU and VPQ proposed in FIERY \cite{fiery} for the motion prediction task, our proposed APA (Accident Prediction Accuracy) and id error, position error for accident prediction task, and detection mAP averaged over center distance matching thresholds of \{1,2,4\} meters for 3D object detection task.

\begin{table}[t!]
    \begin{center}
    \small
    \setlength{\tabcolsep}{0pt}
        \begin{tabular}{c|c c|c|c}
            \toprule[0.3mm]
            \multirow{2}{*}{Config} & \multicolumn{2}{c|}{Motion} & \hspace{1em}Accident\hspace{1em} & \hspace{0.5em}{Detection}\\

            & mIOU($\scriptstyle \uparrow$) & VPQ($\scriptstyle \uparrow$) & \hspace{0.5em}APA($\scriptstyle \uparrow$) & \hspace{0.5em}mAP($\scriptstyle \uparrow$)\\
            \midrule
            Average Fusion & 52.1 & 39.5 & 67.1 & 36.2\\
            DiscoNet & 54.2 & 42.0 & 68.9 & 38.5\\
            V2X-ViT & 55.1 & 43.2 & 69.1 & 40.1\\
            CoBEVT & \textbf{56.2} & \textbf{44.0} & \textbf{69.5} & \textbf{40.8}\\
            \bottomrule[0.3mm]
        \end{tabular}
    \caption{V2XFormer with different V2X fusion modules including the average pooling baseline and state-of-the art methods under five V2X agents setting.} 
    \label{tab:experiment_v2x_fusion}
    \end{center}
\end{table}

\begin{table}[t!]
    \begin{center}
    \small
    \setlength{\tabcolsep}{0pt}
        \begin{tabular}{c|c c|c c c|c}
            \toprule[0.3mm]
            \multirow{2}{*}{Config} & \multicolumn{2}{c|}{Motion} & \multicolumn{3}{c|}{Accident} & \multirow{2}{*}{mAP}\\

            & mIOU($\scriptstyle \uparrow$) & VPQ($\scriptstyle \uparrow$) & APA($\scriptstyle \uparrow$) & id err($\scriptstyle \downarrow$) & pos err($\scriptstyle \downarrow$) & ($\scriptstyle \uparrow$)\\
            \midrule
            Single vehicle & 43.8 & 31.6 & 61.9 & 0.12 & 3.20 & 26.5\\
            ego+behind vehicle& 51.3 & 39.2 & 66.8 & 0.11 & 2.87 & 36.3\\
            ego+other vehicle & 52.1 & 39.9 & 67.4 & 0.10 & 2.85 & 36.6\\
            ego+infra & 52.7 & 40.1 & 68.1 & 0.10 & 2.80 & 36.8\\
            ego+behind+other & 53.6 & 41.2 & 68.4 & 0.08 & 2.87 & 38.1\\
            4 vehicles & 55.5 & 42.5 & 68.9 & 0.07 & 2.91 & 39.0\\
            4 vehicles+infra & \textbf{56.2} & \textbf{44.0} & \textbf{69.5} & \textbf{0.06} & \textbf{2.45} & \textbf{40.8}\\
            \bottomrule[0.3mm]
        \end{tabular}
    \caption{Performance comparison between the single-vehicle model and different V2X configuration models.} 
    \label{tab:experiment_overall}
    \end{center}
\end{table}

\noindent \textbf{V2X fusion module.} We choose the five-agent V2X setting, and compare the performance of utilizing average pooling baseline and various state-of-the-art V2X fusion modules as V2XFormer's fusion module. The V2X fusion methods used include DiscoNet \cite{DiscoNet}, V2X-ViT \cite{V2XVit}, and CoBEVT \cite{COBEVT}. As shown in Table \ref{tab:experiment_v2x_fusion}, CoBEVT performs the best in all three tasks and we will use this V2X fusion module for the following experiments.

\noindent \textbf{Overall performance.} As shown in Table \ref{tab:experiment_overall}, V2X models significantly outperform the single-vehicle baseline in all three tasks. The V2X model with four vehicles and infrastructure exhibits much better performance than the single-vehicle model, with an increase of 12.4, 7.6, and 14.3 in mIOU, APA, and detection mAP, respectively. For the two agent cases, compared to V2X communication with the vehicle behind the ego vehicle (V2X-behind), V2X-other and V2X-infra both demonstrate better performance in all tasks, possibly due to the enhanced visibility on the other side and the broad visibility provided by the infrastructure's relatively high sensor mounting position. Finally, the gradual incorporation of more V2X agents for communication can lead to gradual improvement in performance across all tasks.

\begin{figure}[t!]
  \centering
  \includegraphics[width=1.0\linewidth]{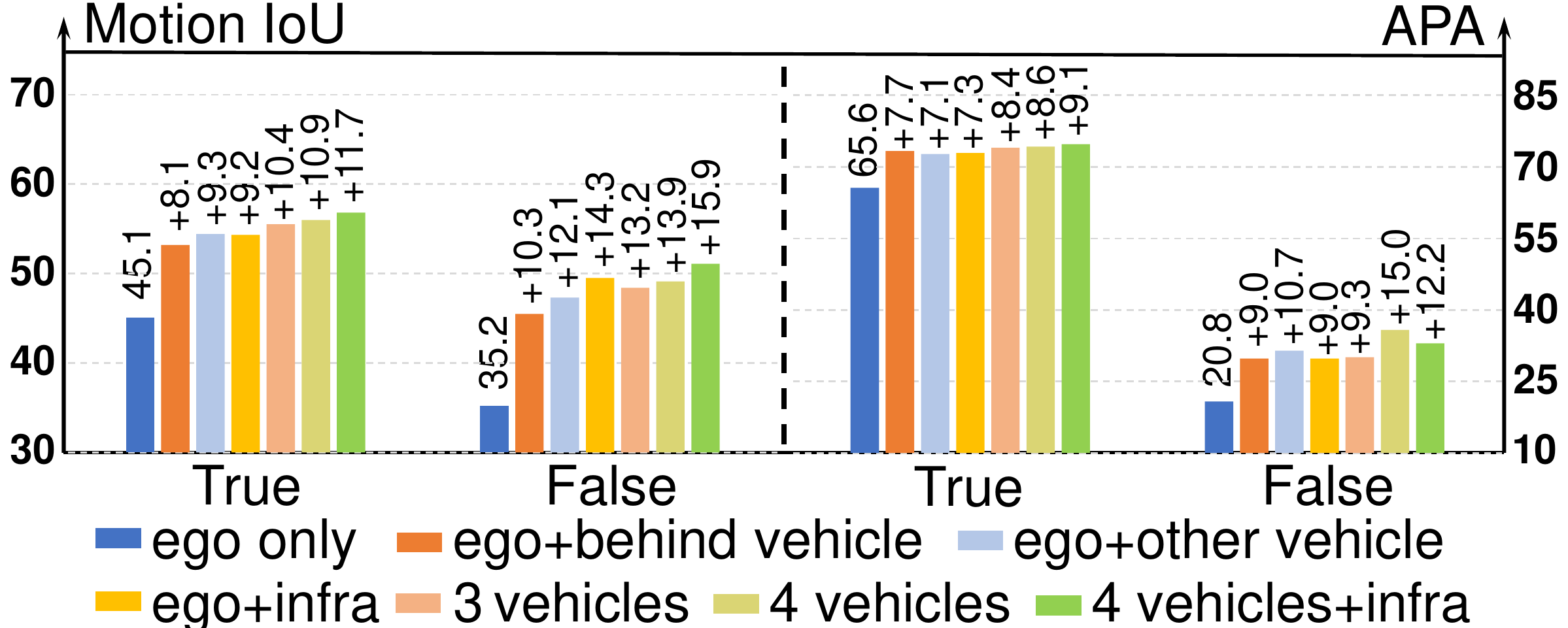}
  \caption{Performance comparison between the single-vehicle model and different v2x configuration models \emph{v.s.} accident visibility.}
  \label{fig:motion_prediction_accident_visibility}
\end{figure}

\begin{table}[t!]
    \small
    \setlength{\tabcolsep}{2pt}
    \begin{center}
        \begin{tabular}{c|c c c c c}
            \hline 
             Prediction horizon & all data & 1s & 2s & 3s & 4s\\
             \hline 
             2s & 61.9 & 74.7 & 28.7 & none & none\\
            3s & 50.5 & 71.5 & 25.7 & 21.2 & none\\
            4s & 35.4 & 56.3 & 20.4 & 14.6 & 10.2\\
            \hline 
        \end{tabular}

    \caption{Performance of single-vehicle models with different prediction horizon settings at different Time-To-Collision (TTC) for APA metric.}
    \label{tab:motion_prediction_longer_prediction_horizon}
    \end{center}
\end{table}

\noindent \textbf{Performance based on accident vehicle visibility.} During the observation period, accident vehicles or pedestrians may be temporarily or consistently invisible from the ego vehicle's perspective, making it challenging to predict their motion with a single-vehicle model and hindering accident prediction. To address this, we evaluate the performance of different V2X models and a single-vehicle model by dividing the evaluation data based on accident vehicle or pedestrian visibility from the ego vehicle side. We define a sample with over half of its observation frames having invisible accident vehicles as an invisible sample for accidents. 
Figure \ref{fig:motion_prediction_accident_visibility} shows that the performance gap between V2X models and the single-vehicle model is significantly larger when there is limited accident visibility from the ego vehicle side, for both motion prediction and accident prediction tasks. Specifically, V2X-5agent model (4 vehicles + infra) outperforms the single-vehicle model by 15.9 and 12.2 higher mIOU and APA, respectively, for invisible accident scenarios, while the gap is only 11.7 and 9.1 in terms of mIOU and APA for visible accident scenarios.

\begin{figure*}[t!]
  \centering
  \includegraphics[width=1.0\textwidth]{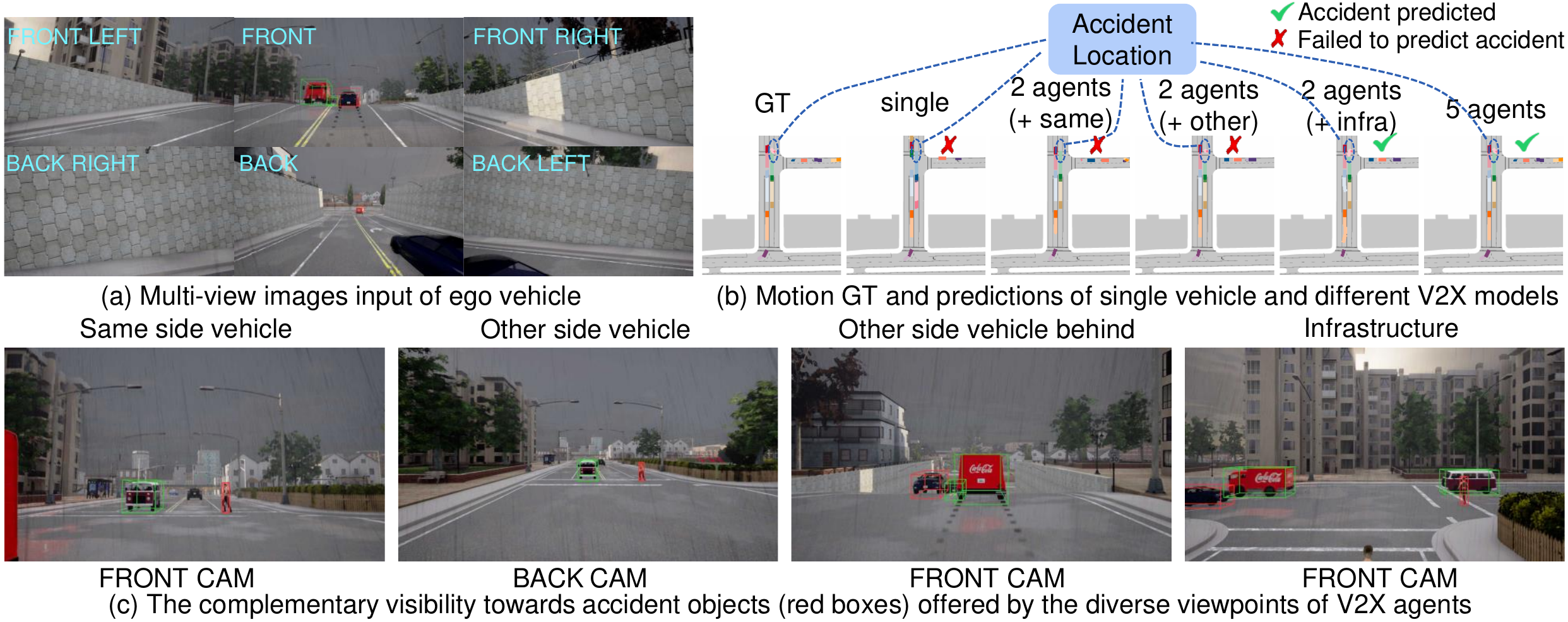}
  \caption{A qualitative result where the ego vehicle is going uphill while the ahead vehicle will collide with the pedestrian crossing the road. The crossing pedestrian is invisible to the ego vehicle due to the uphill terrain. In this case, the infrastructure provides clear visibility for the colliding vehicle and pedestrian, thus successfully predicting the accidents for V2X-infra and V2X-5agent model.}
  \label{fig:qualitative_results}
\end{figure*}
\noindent \textbf{Longer prediction horizon.} We also conduct experiments on predicting longer future motion and choose the single-vehicle model as the baseline. As shown in Table \ref{tab:motion_prediction_longer_prediction_horizon}, predicting longer future motion achieves worse accident prediction accuracy than the model with a shorter prediction horizon. For example, the model predicting 4s future achieves almost half the APA of the 2s-setting model on all validation data, achieving 35.4 and 61.9, respectively. On the other hand, the 4s-setting model achieves an APA of 10.2 for samples 4s prior to the collision, while other models are unable to predict the accident this early due to their design. These results suggest a trade-off between predicting longer future horizons and achieving satisfactory overall performance.

\begin{figure}[t!]
  \centering
  \includegraphics[width=1.0\linewidth]{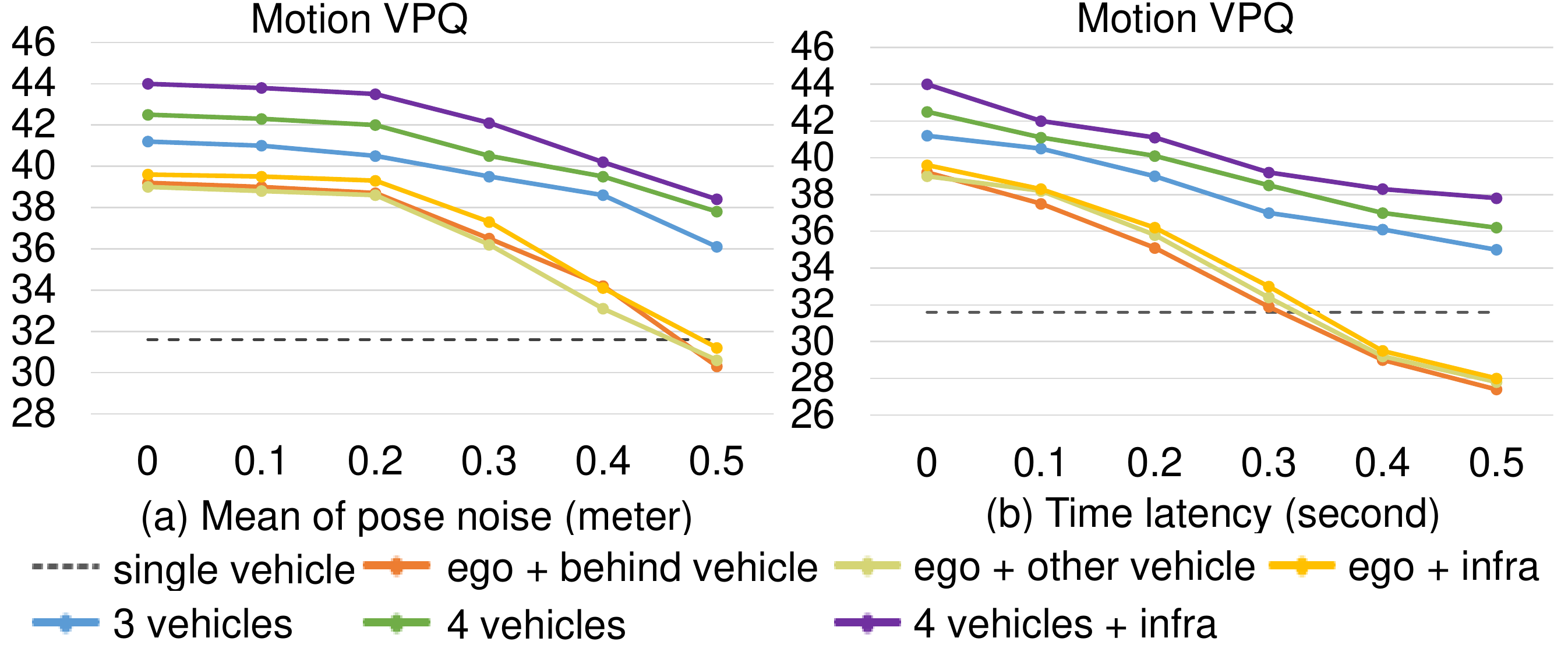}
  \caption{Robustness test on pose error and latency.}
  \label{fig:robustness test}
\end{figure}
\noindent \textbf{Qualitative results.} Figure \ref{fig:qualitative_results} shows an example where the crossing pedestrian is invisible to the ego vehicle due to the uphill terrain. As a result, the single-vehicle model is unable to detect the pedestrian and fails to predict the upcoming accident. In contrast, the infrastructure provides complementary visibility for the colliding vehicle and pedestrian, allowing the V2X-infra and V2X-5agents models to accurately anticipate the accident.

\noindent \textbf{Robustness on pose error and latency.}
We test the robustness of V2XFormer with different V2X configurations against pose noise (Gaussian noise with a mean from 0.1m to 0.5m and a standard deviation of 0.02m), and delay latency (0.1s to 0.5s). As shown in Figure \ref{fig:robustness test}, incorporating more V2X agents provides stronger robustness against pose error and communication latency. 
Nevertheless, communicating with more agents results in larger data transmission, which can subsequently increase time latency and hinder performance. Therefore, a trade-off is necessary when considering V2X configurations.

\noindent \textbf{Sim2Real Domain Adaptation.}
To validate the real-world generalization ability of the models trained with our proposed DeepAccident dataset, we fine-tune the trained single vehicle model on nuScenes for five epochs and compare it with the original BEVerse-tiny model that is only trained on nuScenes for motion prediction and 3D object detection tasks. As shown in Table \ref{tab:nuscenes_adptation}, the model trained with both datasets achieves 1.9 higher mAP and 0.8 higher VPQ on the nuScenes validation dataset, indicating the usefulness of our proposed DeepAccident dataset for real-world scenarios.

\begin{table}[t!]
    \small
    \renewcommand{\tabcolsep}{1mm}
    \begin{center}
        \begin{tabular}{c|c c}
            \hline
            Training data & VPQ & mAP\\
            \hline 
              nuScenes only & 33.4 & 32.1\\
              DeepAccident + nuScenes & \textbf{34.2 (+0.8)} & \textbf{34.0 (+1.9)}\\
            \hline 
        \end{tabular}
    \caption{Performance comparison between the original BEVerse-tiny model and the model trained with both the synthesized DeepAccident data and the real-world nuScenes data for motion prediction and 3D object detection.}
    \label{tab:nuscenes_adptation}
    \end{center}
\end{table}

\section{Conclusion}
We propose DeepAccident, the first large-scale V2X autonomous driving dataset that includes various collision accident scenarios commonly encountered in real-world driving. Based on this dataset, we introduce the end-to-end motion and accident prediction task and corresponding metrics to assess the accuracy of accident prediction. DeepAccident contains sensor data and annotation labels from four vehicles and one infrastructure for each scenario, allowing for the V2X research for perception and prediction. Our proposed V2XFormer outperforms the single-vehicle model in both perception and prediction tasks, providing a baseline for future research. The proposed DeepAccident serves as a direct safety benchmark for autonomous driving algorithms and as a supplementary dataset for both single-vehicle and V2X perception research in safety-critical scenarios.

\noindent \textbf{Acknowledgement.} This paper is partially supported by the General Research Fund of Hong Kong No.17200622.

{
\bibliography{aaai24}

\begin{thebibliography}{24}
\providecommand{\natexlab}[1]{#1}

\bibitem[{Caesar et~al.(2020)Caesar, Bankiti, Lang, Vora, Liong, Xu, Krishnan,
  Pan, Baldan, and Beijbom}]{nuscenes2020cvpr}
Caesar, H.; Bankiti, V.; Lang, A.~H.; Vora, S.; Liong, V.~E.; Xu, Q.; Krishnan,
  A.; Pan, Y.; Baldan, G.; and Beijbom, O. 2020.
\newblock nuscenes: A multimodal dataset for autonomous driving.
\newblock In \emph{Proceedings of the IEEE/CVF conference on computer vision
  and pattern recognition}, 11621--11631.

\bibitem[{Dosovitskiy et~al.(2017)Dosovitskiy, Ros, Codevilla, Lopez, and
  Koltun}]{carla}
Dosovitskiy, A.; Ros, G.; Codevilla, F.; Lopez, A.; and Koltun, V. 2017.
\newblock CARLA: An open urban driving simulator. arXiv 2017.
\newblock \emph{arXiv preprint arXiv:1711.03938}.

\bibitem[{Geiger et~al.(2013)Geiger, Lenz, Stiller, and Urtasun}]{KITTI}
Geiger, A.; Lenz, P.; Stiller, C.; and Urtasun, R. 2013.
\newblock Vision meets robotics: The kitti dataset.
\newblock \emph{The International Journal of Robotics Research}, 32(11):
  1231--1237.

\bibitem[{Hoon et~al.(2019)Hoon, Kangwook, Gyeongjo, and
  Changho}]{crash_to_not_crash}
Hoon, K.; Kangwook, L.; Gyeongjo, H.; and Changho, S. 2019.
\newblock Crash to Not Crash: Learn to Identify Dangerous Vehicles Using a
  Simulator.
\newblock In \emph{Proceedings of the AAAI Conference on Artificial
  Intelligence}, 978–985.

\bibitem[{Hu et~al.(2021)Hu, Murez, Mohan, Dudas, Hawke, Badrinarayanan,
  Cipolla, and Kendall}]{fiery}
Hu, A.; Murez, Z.; Mohan, N.; Dudas, S.; Hawke, J.; Badrinarayanan, V.;
  Cipolla, R.; and Kendall, A. 2021.
\newblock FIERY: future instance prediction in bird's-eye view from surround
  monocular cameras.
\newblock In \emph{Proceedings of the IEEE/CVF International Conference on
  Computer Vision}, 15273--15282.

\bibitem[{Huang et~al.(2021)Huang, Huang, Zhu, and Du}]{bevdet}
Huang, J.; Huang, G.; Zhu, Z.; and Du, D. 2021.
\newblock Bevdet: High-performance multi-camera 3d object detection in
  bird-eye-view.
\newblock \emph{arXiv preprint arXiv:2112.11790}.

\bibitem[{Lang et~al.(2019)Lang, Vora, Caesar, Zhou, Yang, and
  Beijbom}]{pointpillar}
Lang, A.~H.; Vora, S.; Caesar, H.; Zhou, L.; Yang, J.; and Beijbom, O. 2019.
\newblock PointPillars: Fast Encoders for Object Detection from Point Clouds.
\newblock In \emph{Proceedings of the IEEE/CVF conference on computer vision
  and pattern recognition}, 12697--12705.

\bibitem[{Li et~al.(2023)Li, Huang, Chen, Cui, Liang, Shen, Liu, Xie, Sheng,
  Ouyang et~al.}]{fasetbev}
Li, Y.; Huang, B.; Chen, Z.; Cui, Y.; Liang, F.; Shen, M.; Liu, F.; Xie, E.;
  Sheng, L.; Ouyang, W.; et~al. 2023.
\newblock Fast-BEV: A Fast and Strong Bird's-Eye View Perception Baseline.
\newblock \emph{arXiv preprint arXiv:2301.12511}.

\bibitem[{Li et~al.(2022)Li, Ma, An, Wang, Zhong, Chen, and Feng}]{v2x-sim}
Li, Y.; Ma, D.; An, Z.; Wang, Z.; Zhong, Y.; Chen, S.; and Feng, C. 2022.
\newblock V2X-Sim: Multi-agent collaborative perception dataset and benchmark
  for autonomous driving.
\newblock \emph{IEEE Robotics and Automation Letters}, 7(4): 10914--10921.

\bibitem[{Li et~al.(2021)Li, Ren, Wu, Chen, Feng, and Zhang}]{DiscoNet}
Li, Y.; Ren, S.; Wu, P.; Chen, S.; Feng, C.; and Zhang, W. 2021.
\newblock Learning Distilled Collaboration Graph for Multi-Agent Perception.
\newblock In \emph{Thirty-fifth Conference on Neural Information Processing
  Systems (NeurIPS 2021)}.

\bibitem[{Liu et~al.(2021)Liu, Lin, Cao, Hu, Wei, Zhang, Lin, and
  Guo}]{swin-transformer}
Liu, Z.; Lin, Y.; Cao, Y.; Hu, H.; Wei, Y.; Zhang, Z.; Lin, S.; and Guo, B.
  2021.
\newblock Swin transformer: Hierarchical vision transformer using shifted
  windows.
\newblock In \emph{Proceedings of the IEEE/CVF international conference on
  computer vision}, 10012--10022.

\bibitem[{Najm et~al.(2007)Najm, Smith, Yanagisawa et~al.}]{pre-crash}
Najm, W.~G.; Smith, J.~D.; Yanagisawa, M.; et~al. 2007.
\newblock Pre-crash scenario typology for crash avoidance research.
\newblock Technical report, United States. National Highway Traffic Safety
  Administration.

\bibitem[{Rempe et~al.(2022)Rempe, Philion, Guibas, Fidler, and
  Litany}]{STRIVE}
Rempe, D.; Philion, J.; Guibas, L.~J.; Fidler, S.; and Litany, O. 2022.
\newblock Generating Useful Accident-Prone Driving Scenarios via a Learned
  Traffic Prior.
\newblock In \emph{Conference on Computer Vision and Pattern Recognition
  (CVPR)}.

\bibitem[{Sadegh~Aliakbarian et~al.(2018)Sadegh~Aliakbarian, Sadat~Saleh,
  Salzmann, Fernando, Petersson, and Andersson}]{VIENA2}
Sadegh~Aliakbarian, M.; Sadat~Saleh, F.; Salzmann, M.; Fernando, B.; Petersson,
  L.; and Andersson, L. 2018.
\newblock VIENA2: A Driving Anticipation Dataset.
\newblock \emph{arXiv e-prints}, arXiv--1810.

\bibitem[{Shi et~al.(2020)Shi, Guo, Jiang, Wang, Shi, Wang, and Li}]{pvrcnn}
Shi, S.; Guo, C.; Jiang, L.; Wang, Z.; Shi, J.; Wang, X.; and Li, H. 2020.
\newblock PV-RCNN: Point-Voxel Feature Set Abstraction for 3D Object Detection.
\newblock In \emph{Proceedings of the IEEE/CVF Conference on Computer Vision
  and Pattern Recognition}, 10529--10538.

\bibitem[{Sun et~al.(2020)Sun, Kretzschmar, Dotiwalla, Chouard, Patnaik, Tsui,
  Guo, Zhou, Chai, Caine et~al.}]{Waymo}
Sun, P.; Kretzschmar, H.; Dotiwalla, X.; Chouard, A.; Patnaik, V.; Tsui, P.;
  Guo, J.; Zhou, Y.; Chai, Y.; Caine, B.; et~al. 2020.
\newblock Scalability in perception for autonomous driving: Waymo open dataset.
\newblock In \emph{Proceedings of the IEEE/CVF conference on computer vision
  and pattern recognition}, 2446--2454.

\bibitem[{Wang et~al.(2021)Wang, Pun, Tu, Manivasagam, Sadat, Casas, Ren, and
  Urtasun}]{AdvSim}
Wang, J.; Pun, A.; Tu, J.; Manivasagam, S.; Sadat, A.; Casas, S.; Ren, M.; and
  Urtasun, R. 2021.
\newblock AdvSim: Generating Safety-Critical Scenarios for Self-Driving
  Vehicles.
\newblock \emph{Conference on Computer Vision and Pattern Recognition (CVPR)}.

\bibitem[{Xu et~al.(2022{\natexlab{a}})Xu, Tu, Xiang, Shao, Bolei, and
  Ma}]{COBEVT}
Xu, R.; Tu, Z.; Xiang, H.; Shao, W.; Bolei, Z.; and Ma, J. 2022{\natexlab{a}}.
\newblock CoBEVT: Cooperative Bird's Eye View Semantic Segmentation with Sparse
  Transformers.
\newblock In \emph{Conference on Robot Learning (CoRL)}.

\bibitem[{Xu et~al.(2022{\natexlab{b}})Xu, Xiang, Tu, Xia, Yang, and
  Ma}]{V2XVit}
Xu, R.; Xiang, H.; Tu, Z.; Xia, X.; Yang, M.-H.; and Ma, J. 2022{\natexlab{b}}.
\newblock V2X-ViT: Vehicle-to-Everything Cooperative Perception with Vision
  Transformer.
\newblock In \emph{Proceedings of the European Conference on Computer Vision
  (ECCV)}.

\bibitem[{Xu et~al.(2022{\natexlab{c}})Xu, Xiang, Xia, Han, Li, and Ma}]{OPV2V}
Xu, R.; Xiang, H.; Xia, X.; Han, X.; Li, J.; and Ma, J. 2022{\natexlab{c}}.
\newblock Opv2v: An open benchmark dataset and fusion pipeline for perception
  with vehicle-to-vehicle communication.
\newblock In \emph{2022 International Conference on Robotics and Automation
  (ICRA)}, 2583--2589. IEEE.

\bibitem[{Xu et~al.(2022{\natexlab{d}})Xu, Huang, Nan, and Lian}]{TAD}
Xu, Y.; Huang, C.; Nan, Y.; and Lian, S. 2022{\natexlab{d}}.
\newblock TAD: A Large-Scale Benchmark for Traffic Accidents Detection from
  Video Surveillance.
\newblock \emph{arXiv preprint arXiv:2209.12386}.

\bibitem[{Yu et~al.(2022)Yu, Luo, Shu, Huo, Yang, Shi, Guo, Li, Hu, Yuan, and
  Nie}]{DAIR-V2X}
Yu, H.; Luo, Y.; Shu, M.; Huo, Y.; Yang, Z.; Shi, Y.; Guo, Z.; Li, H.; Hu, X.;
  Yuan, J.; and Nie, Z. 2022.
\newblock Dair-v2x: A large-scale dataset for vehicle-infrastructure
  cooperative 3d object detection.
\newblock In \emph{Proceedings of the IEEE/CVF Conference on Computer Vision
  and Pattern Recognition}, 21361--21370.

\bibitem[{Yu et~al.(2023)Yu, Yang, Ruan, Yang, Tang, Gao, Hao, Shi, Pan, Sun,
  Song, Yuan, Luo, and Nie}]{V2X-seq}
Yu, H.; Yang, W.; Ruan, H.; Yang, Z.; Tang, Y.; Gao, X.; Hao, X.; Shi, Y.; Pan,
  Y.; Sun, N.; Song, J.; Yuan, J.; Luo, P.; and Nie, Z. 2023.
\newblock V2X-Seq: A large-scale sequential dataset for vehicle-infrastructure
  cooperative perception and forecasting.
\newblock In \emph{Proceedings of the IEEE/CVF Conference on Computer Vision
  and Pattern Recognition}.

\bibitem[{Zhang et~al.(2022)Zhang, Zhu, Zheng, Huang, Huang, Zhou, and
  Lu}]{beverse}
Zhang, Y.; Zhu, Z.; Zheng, W.; Huang, J.; Huang, G.; Zhou, J.; and Lu, J. 2022.
\newblock Beverse: Unified perception and prediction in birds-eye-view for
  vision-centric autonomous driving.
\newblock \emph{arXiv preprint arXiv:2205.09743}.

\end{thebibliography}
}

\newpage
\appendix

\section{Ablation Experiments}
\begin{table}[t!]
    \begin{center}
    \small
    \setlength{\tabcolsep}{1.5pt}
        \begin{tabular}{c|c c c|c c c}
            \toprule[0.3mm]
            \multirow{2}{*}{Config} & \multicolumn{3}{c|}{Validation} & \multicolumn{3}{c}{Testing}\\

            & VPQ & APA & mAP & VPQ & APA & mAP\\
            \midrule
            Single vehicle & 31.6 & 61.9 & 26.5 & 30.5 & 61.9 & 25.8\\
            + behind vehicle & 39.2 & 66.8 & 36.3 & 38.9 & 65.8 & 36.0\\
            + other vehicle & 39.0 & 67.2 & 36.5 & 38.7 & 66.2 & 36.1\\
            + infrastructure & 39.6 & 68.1 & 36.8 & 39.1 & 67.3 & 36.2\\
            + behind \& other vehicle & 41.2 & 68.4 & 38.1 & 40.5 & 68.0 & 37.8\\
            4 vehicles & 42.5 & 68.9 & 39.0 & 41.8 & 68.0 & 38.3\\
            4 vehicles + infra & \textbf{44.0} & \textbf{69.5} & \textbf{40.8} & \textbf{43.2} & \textbf{68.7} & \textbf{39.8}\\
            \bottomrule[0.3mm]
        \end{tabular}
    \caption{Performance comparison between single-vehicle model and different v2x configuration models on validation and testing split of DeepAccident. 
    } 
    \label{tab:experiment_val_vs_test}
    \end{center}
\end{table}

\noindent \textbf{Performance comparison on val and test split.} 
In the main paper, we evaluated all experiment results on the validation split of DeepAccident. Table \ref{tab:experiment_val_vs_test} presents a performance comparison of various trained models on both the validation and testing splits. The table shows that the models perform similarly on both splits, with consistent relative performance gaps between models. For example, the V2X model with four vehicles and one infrastructure outperforms the single-vehicle model by 12.4 and 12.7 higher VPQ on the validation and testing splits, achieving 44.0 and 43.2 VPQ, respectively. These results confirm the validity of the split in DeepAccident.

\begin{figure*}[t!]
  \centering
  \includegraphics[width=1.0\textwidth]{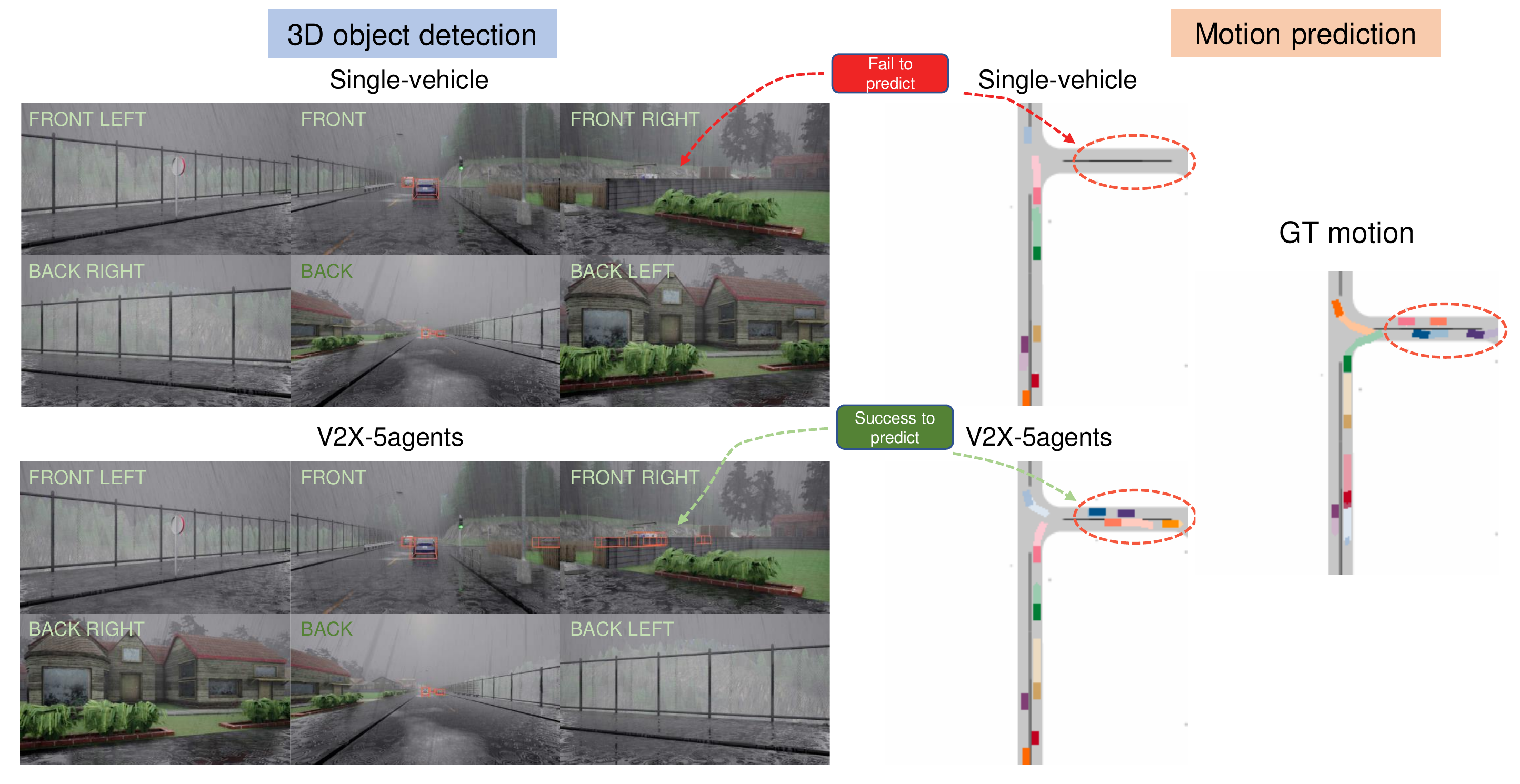}
  \caption{Qualitative comparison between the single-vehicle model and the V2X-5agents model on 3D object detection and motion prediction tasks. The red circles on motion figures indicate the front right area behind a wall such that the single-vehicle model failed to predict objects and their motions in that area while the V2X-5agents model succeeded.}
  \label{fig:qualitative_visualization_motion_detection}
  \vspace{-1em}
\end{figure*}

\begin{figure*}[t!]
  \centering
  \includegraphics[width=0.95\textwidth]{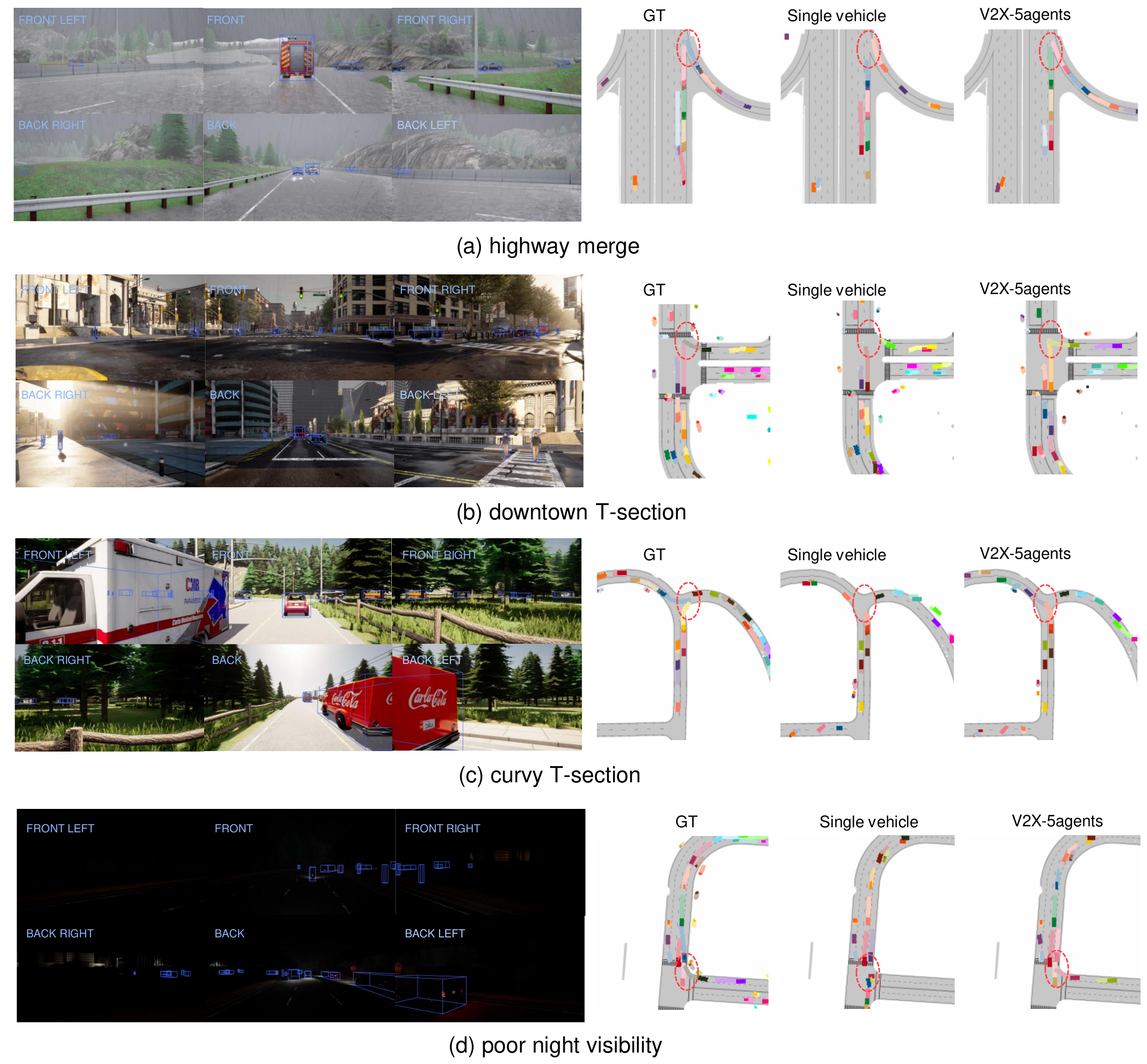}
  \caption{Qualitative comparison between the single-vehicle model and the V2X-5agents model across various scenarios on end-to-end motion and accident detection tasks. The areas in the motion prediction results marked by red circles indicate the happening place of accidents. The V2X-5agents model shows superiority over the single-vehicle model for both motion and accident prediction.}
  \label{fig:qualitative_visualization}
  \vspace{-1.5em}
\end{figure*}

\noindent \textbf{Qualitative results.} 
Figures \ref{fig:qualitative_visualization_motion_detection} and \ref{fig:qualitative_visualization} present additional qualitative results comparing the V2X-5agents model and the single-vehicle model. These figures demonstrate the improved prediction and perception capabilities of the V2X model over the single-vehicle baseline, particularly in scenarios where the ego vehicle's visibility is reduced due to occlusion, weather, or lighting conditions.

\begin{table}[t!]
    \renewcommand{\tabcolsep}{1mm}
    \small
    \begin{center}
        \begin{tabular}{p{2cm}|p{5.5cm}}
            \toprule
            Sensor & Description \\
            \midrule 
            Vehicles &   For each vehicle:
            \begin{enumerate}[itemsep=-0.3em, label=•, leftmargin=0.3cm]
   \item Six rgb cameras with 1600 x 900 resolution. $70^{\circ}$ horizontal FoV for F/FL/FR/BL/BR cameras. $110^{\circ}$ horizontal FoV for the Back camera.
   \item One 32-lane LiDAR mounted on top with 70 meter range, 30° vertical FoV, and 10 Hz rotation frequency. 
\end{enumerate}\\
            
            \midrule 
            Infrastructure & 
            \begin{enumerate}[itemsep=-0.3em, label=•, leftmargin=0.3cm]
   \item Same senor configurations as the vehicles - six rgb cameras and one LiDAR.
   \item Sensors installed on a high position randomly from 3 to 5 meters. 
   \end{enumerate} \\
            
            \hline 
            V2X communication &  
            \begin{enumerate}[itemsep=-0.3em, label=•, leftmargin=0.3cm]
   \item Recorded relative poses between different V2X agents to align the information from different sources. 
   \end{enumerate}\\
            \midrule
        \end{tabular}
    \caption{Sensor configurations in DeepAccident. As for the V2X setting, we set four vehicles and one infrastructure to record data for each scenario.}
    \label{tab:sensor_config}
    \end{center}
\end{table}

\begin{figure*}[t!]
  \centering
  \includegraphics[width=0.95\textwidth]{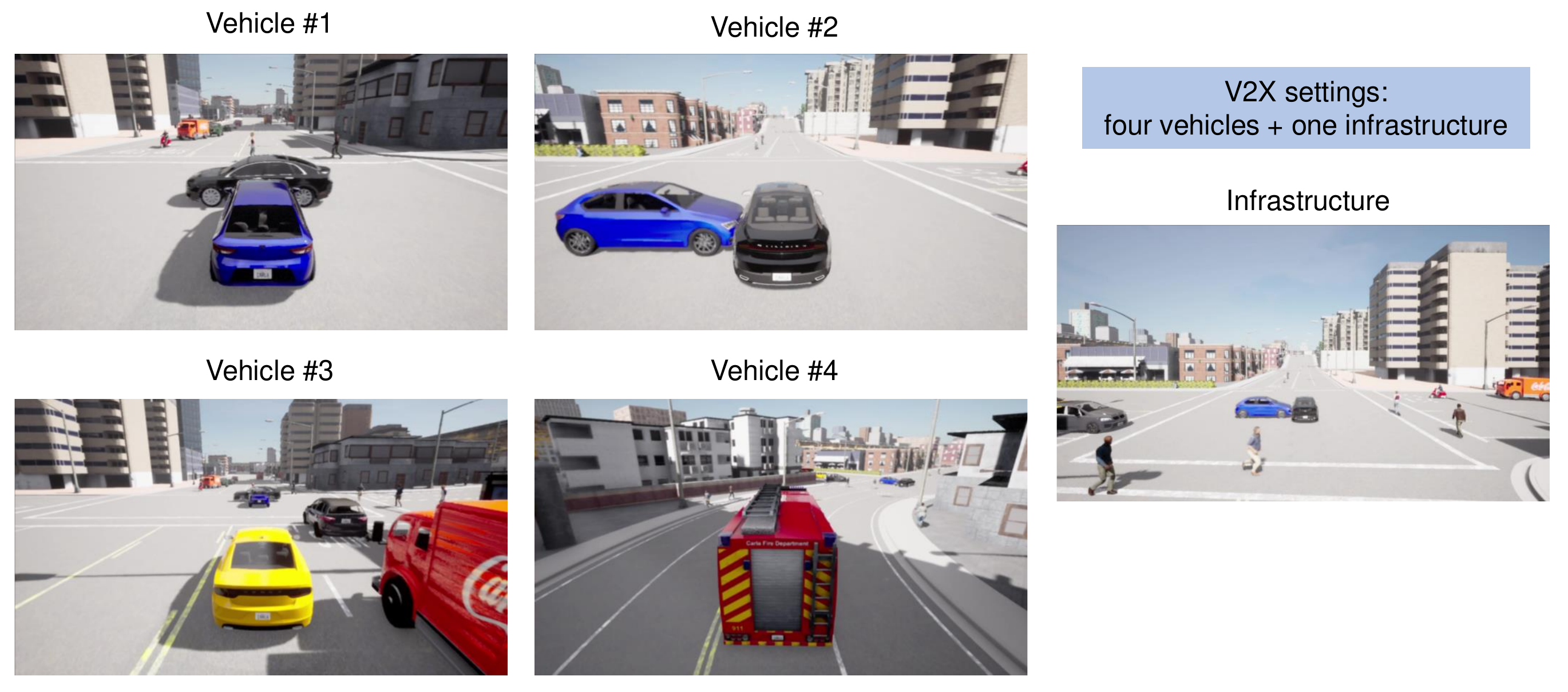}
  \caption{The V2X setting in the proposed DeepAccident dataset. We have four vehicles and one infrastructure to capture the full set of sensor data for each scenario. Vehicle \# 1 and vehicle \# 2 are designed to collide with each other, and vehicle \# 3 and vehicle \# 4 are the V2X vehicles that we designed to follow behind those two accident vehicles. Moreover, the views from the infrastructure are also collected, which often have little occlusion and have better visibility. The images shown here are captured from the cameras mounted behind the vehicles. These images are only for V2X illustration purposes and differ from the captured camera images in the proposed DeepAccident dataset.}
  \label{fig:V2X setting iilustration}
  \vspace{-1em}
\end{figure*}

\section{The DeepAccident Dataset}
\subsection{Scenario Design}
\noindent \textbf{V2X setting illustration.} As illustrated in Figure \ref{fig:V2X setting iilustration}, for each scenario in our proposed DeepAccident dataset, we have four vehicles and one infrastructure to collect sensor data, thus providing diverse viewpoints and enabling V2X research.

\subsection{Dataset Statistics}
\noindent \textbf{Dataset size \& split.} 
While the simulator-based nature of the dataset could have enabled the generation of an unlimited amount of data, we chose to include data on 691 scenarios to maintain a manageable dataset size of roughly 310 GB. 
In addition, our proposed DeepAccident dataset provides six types of annotated objects, including car, van, truck, motorcycle, cyclist, and pedestrian, and supports numerous perception and prediction tasks. 
Moreover, we have meticulously divided the DeepAccident dataset into training, validation, and testing splits to make sure all of them contain enough diverse scenarios, such as scenarios with collision accidents and adverse weather and lighting conditions. With a total of 691 scenarios, we allocated the splits with a ratio of 0.7, 0.15, and 0.15 for training, validation, and testing. As such, the training, validation, and testing splits consist of 483, 104, and 104 scenarios, respectively. As a result, the 285k annotated samples were distributed with 203k, 41k, and 41k samples assigned to the training, validation, and testing splits correspondingly.

\begin{figure*}[t!]
  \centering
  \includegraphics[width=0.85\textwidth]{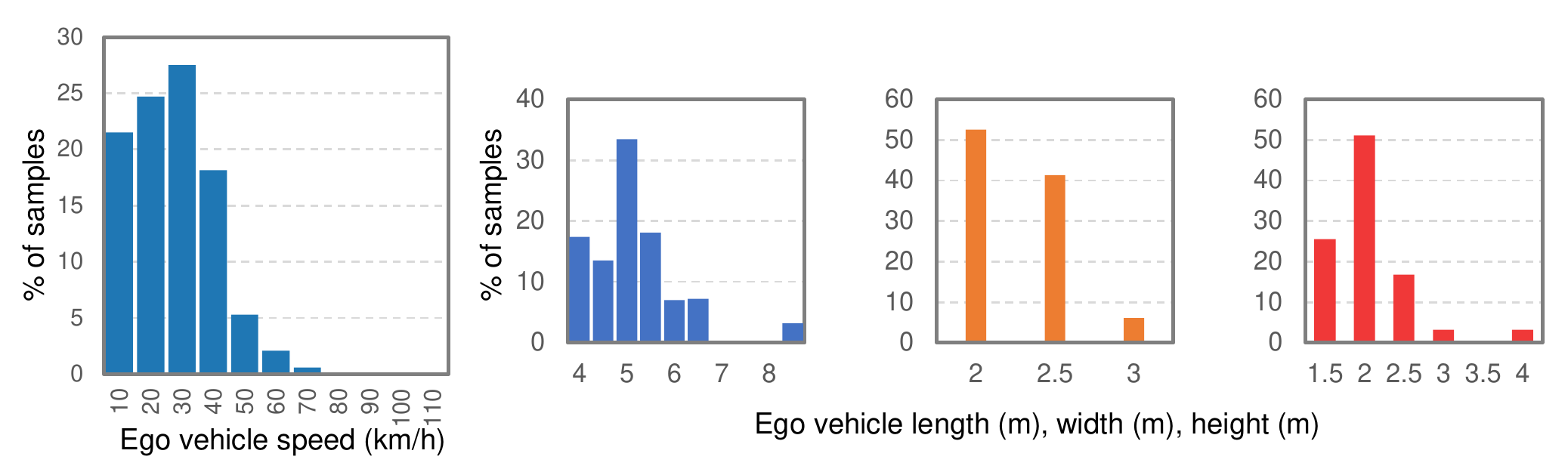}
  \caption{Ego vehicle statistics of the proposed DeepAccident dataset, including the driving speed and size dimensions.}
  \label{fig:ego vehicle statistics}
  \vspace{-2mm}
\end{figure*}

\noindent \textbf{Ego vehicle statistics.} Figure \ref{fig:ego vehicle statistics} shows the ego vehicle statistics of the DeepAccident dataset, including the driving speed and size dimensions. The driving speed of ego vehicle covers a wide range, including high-speed driving cases. Additionally, the ego vehicle size also varies significantly from small cars to huge trucks to offer more viewpoint diversity.

\begin{figure*}[t!]
  \centering
  \includegraphics[width=0.8\textwidth]{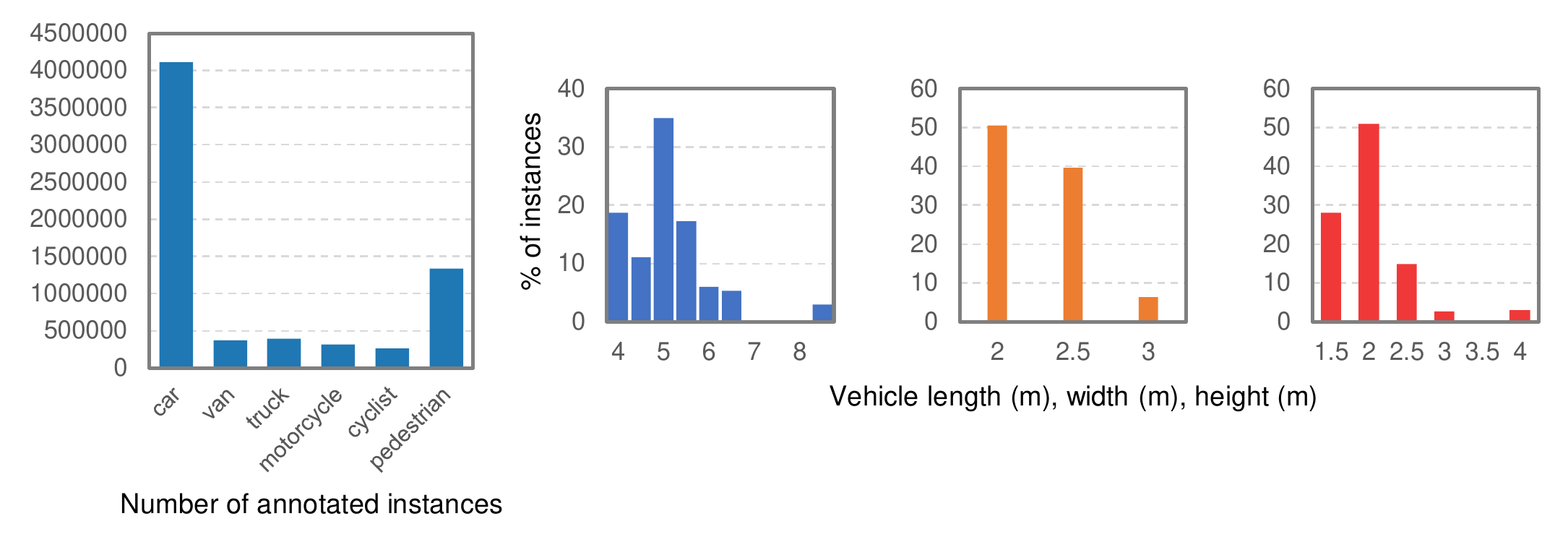}
  \caption{Instance-level statistics, including the number of annotated instances for different classes and the size distribution of vehicle classes (car, van, truck). Each class has a considerable amount of training data, and the vehicles annotated in our DeepAccident dataset show great diversity in their sizes.}
  \label{fig:instance-level statistics}
\end{figure*}

\begin{figure*}[t!]
  \centering
  \includegraphics[width=1.0\textwidth]{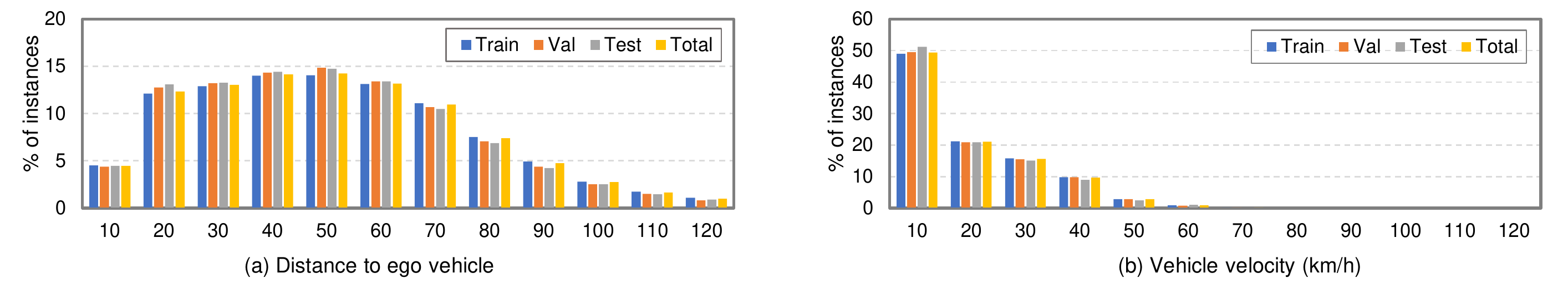}
  \caption{Instance-level vehicle statistics include (a) their distance to the ego vehicle and (b) their velocity. Both figures show the diversity of the vehicles in the DeepAccident dataset. Moreover, the distributions on training, validation, and testing splits show remarkable consistency with the overall data, which shows the validity of our dataset split.}
  \label{fig:instance-level statistics over splits}
\end{figure*}

\noindent \textbf{Instance-level statistics.} Figure \ref{fig:instance-level statistics} and \ref{fig:instance-level statistics over splits} provide instance-level statistics of the DeepAccident dataset. These figures demonstrate that all six object classes in DeepAccident have substantial amounts of annotated data, while the vehicle classes (car, van, and truck) exhibit a wide range of sizes. Moreover, the speed and distance of surrounding vehicles from the ego vehicle also exhibit significant variation. These results show the instance-level diversity of the proposed DeepAccident dataset.

\begin{figure*}[t]
  \centering
  \includegraphics[width=1.0\textwidth]{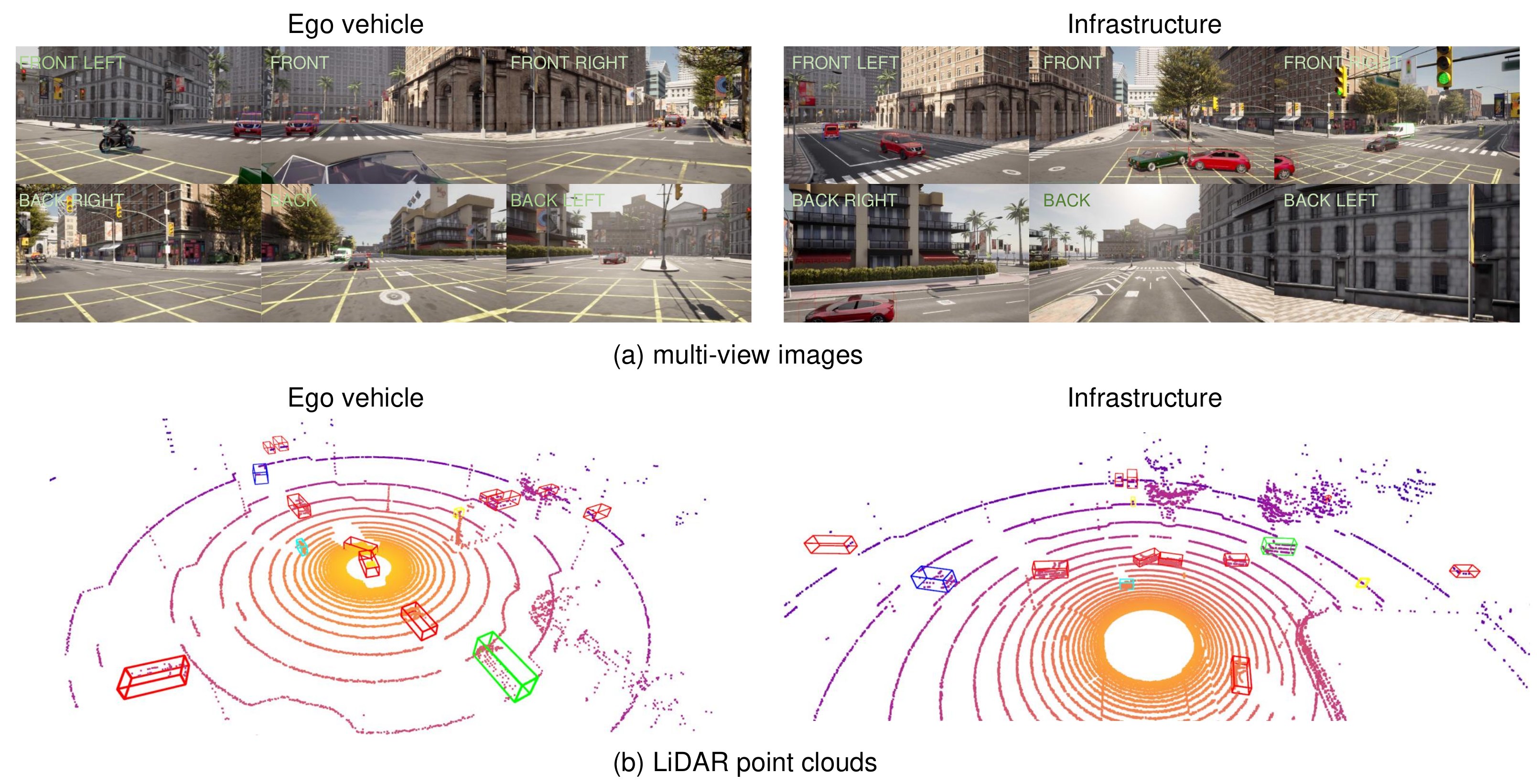}
  \caption{Visualization for the collected multi-view camera images and LiDAR point clouds from the vehicle's and the infrastructure's perspectives when the collision happens. Different box colors indicate different classes.}
  \label{fig:sensor visualization}
\end{figure*}

\begin{figure*}[t!]
  \centering
  \includegraphics[width=1.0\textwidth]{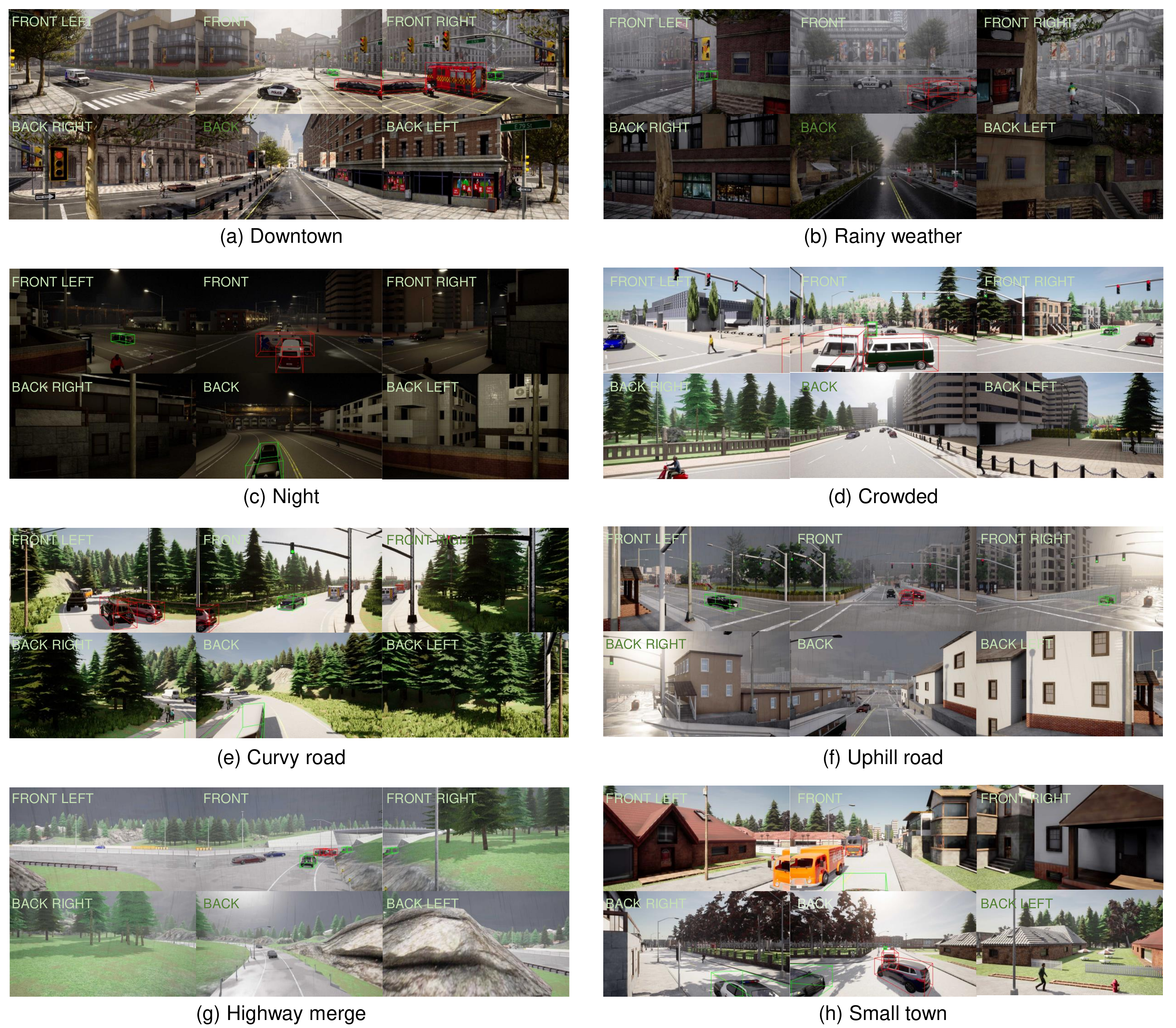}
  \caption{Some showcase accident scenarios in our proposed DeepAccident across different road types, weathers, and time-of-day conditions. All these multi-view camera images are captured from the infrastructure side to better illustrate the scenes. The red bounding boxes in each scenario represent the colliding vehicles, and the green bounding boxes represent the additional V2X vehicles that also collect data.}
  \label{fig:accidents showcase}
\end{figure*}

\noindent \textbf{Dataset visualization.} Additionally, we provide visualization for the collected camera and LiDAR data in Figure \ref{fig:sensor visualization} and some showcase accident scenarios captured by infrastructure in Figure \ref{fig:accidents showcase}. 

\end{document}